\documentclass[journal]{IEEEtran}

\ifCLASSINFOpdf
\else
\fi
\usepackage[numbers]{natbib}
\usepackage{url}
\usepackage{graphicx}
\usepackage{color, soul}
\usepackage{multirow}
\usepackage{hyperref}

\usepackage{easyReview}
\def\remove#1{}
\renewcommand{\add}[1]{\textcolor{black}{#1}}

\hypersetup{
    colorlinks=true,
    linkcolor=magenta,
    citecolor=blue,
    filecolor=magenta,      
    urlcolor=blue,
    pdftitle={Overleaf Example},
    pdfpagemode=FullScreen,
}

\usepackage{booktabs}

\begin{document}

%
\thispagestyle{empty}
{\noindent\Large IEEE Copyright Notice}\\[1pt]

{\noindent Copyright (c) 2023 IEEE

\noindent This work is subject to copyright. All rights are reserved by the Publisher, whether the whole or part of the material is concerned, specifically the rights of translation, reprinting, reuse of illustrations, recitation,broadcasting, reproduction on microfilms or in
any other physical way, and transmission or information storage and retrieval, electronic adaptation, computer software, or by similar or dissimilar methodology now known or hereafter developed.}\\[1em]

{\noindent\large Accepted to be published in: IEEE Geoscience and Remote Sensing Letters, 2023.}\\[1in]

{\noindent Cite as:}\\[1pt]

{\setlength{\fboxrule}{1pt}
 \fbox{\parbox{0.95\textwidth}{E. Nascimento, J. Just, J. Almeida, and T. Almeida, ``Productive Crop Field Detection: A New Dataset and Deep Learning Benchmark Results,'' in \emph{IEEE Geoscience and Remote Sensing Letters, 2023.}}}\\[1in] 
}
{\noindent BibTeX:}\\[1pt]

{\setlength{\fboxrule}{1pt}
 \fbox{\parbox{0.95\textwidth}{
 @InProceedings\{SIBGRAPI 2023,

 \begin{tabular}{p{2mm}lcl}
  & author    & = & \{E. \{Nascimento\} and
               J. \{Just\} and\\ & & &
               J. \{Almeida\} and
               T. \{Almeida\}\},\\

  & title     & = & \{Productive Crop Field Detection: A New Dataset and Deep Learning Benchmark Results\},\\


  & booktitle & = & \{IEEE Geoscience and Remote Sensing Letters\},\\



  & year      & = & \{2023\},\\

  & publisher & = & \{\{IEEE\}\},\\

  \end{tabular}

\}
 }}} 

\clearpage

%
\title{Productive Crop Field Detection: A New Dataset and Deep Learning Benchmark Results}
%
%
%
\author {{
        Eduardo~Nascimento,~
        John~Just,~
        Jurandy~Almeida,~
        and~Tiago~Almeida
}
\thanks{E. Nascimento, T. Almeida and J. Almeida are with the Department of Computer Science, Federal University of São Carlos, São Paulo, Brazil (e-mail:
egnascimento@gmail.com; talmeida@ufscar.br; jurandy.almeida@ufscar.br)}
\thanks{J. Just is with the Department of Data Science and Analytics, John Deere, Iowa, USA (e-mail:johnpjust@gmail.com)}
\thanks{Manuscript received May 18, 2023; revised \replace{September 17, 2014}{July 6, 2023}.}}

%
%

\markboth{IEEE Geoscience and Remote Sensing Letters}%
{Shell \MakeLowercase{\textit{et al.}}: Bare Demo of IEEEtran.cls for Journals}
%

\IEEEoverridecommandlockouts
\IEEEpubid{\makebox[\columnwidth]{978-1-5386-5541-2/18/\$31.00~\copyright2018 IEEE \hfill} \hspace{\columnsep}\makebox[\columnwidth]{ }}
\maketitle
\IEEEpubidadjcol

\begin{abstract}
In precision agriculture, detecting productive crop fields is an essential practice that allows the farmer to evaluate operating performance separately and compare different seed varieties, pesticides, and fertilizers. \replace{However, manually identifying productive fields is often a time-consuming, and error-prone task. Previous studies explore different methods to detect crop fields using advanced machine learning algorithms to bring significant improvements in both aspects, but they often lack good quality labeled data. }{However, manually identifying productive fields is often time-consuming, costly, and subjective. Previous studies explore different methods to detect crop fields using advanced machine learning algorithms to support the specialists' decisions, but they often lack good quality labeled data.} In this context, we propose a high-quality dataset generated by machine operation combined with Sentinel-2 images tracked over time. As far as we know, it is the first one to overcome the lack of labeled samples by using this technique. In sequence, we apply a semi-supervised classification of unlabeled data and state-of-the-art supervised and self-supervised deep learning methods to detect productive crop fields automatically. Finally, the results demonstrate high accuracy in Positive Unlabeled learning, which perfectly fits the problem where we have high confidence in the positive samples. Best performances have been found in Triplet Loss Siamese given the existence of an accurate dataset and Contrastive Learning considering situations where we do not have a comprehensive labeled dataset available.
\end{abstract}


%
\IEEEpeerreviewmaketitle

\section{Introduction}
%
%
%
%

\IEEEPARstart{F}{ood} production needs to grow by 70\% to meet the demands of the expected world population by 2050~\cite{nelson2010}. Motivated by this challenge, agriculture has adopted technologies to improve and optimize input returns while preserving natural resources. Integrating these technologies promotes a farming management concept known as \textit{precision agriculture}~\cite{zhang2002}. The main goal of precision agriculture is to provide tools for allowing the farmer to observe, measure, and respond to field variability in crops facilitating faster and better decisions. In addition, these techniques are generic enough to be applied to various crops, including but not limited to corn, soy, coffee, sugarcane, beans, and even pastures~\cite{mulla2013,bhakta2019}.

To efficiently organize and manage large crops, farmers leverage remote sensing to divide their land into smaller observation units that this work will refer to as \textit{agricultural} or \textit{crop fields}. A field shape is designed based on topography and mechanization planning. For example, fields are built around contour farming and the ideal length to fill the wagon capacity in a sugarcane crop. The same logic is followed for grains in the Brazilian South and Southeast. Nonetheless, the most important variable is the harvester capacity in the Midwest, where the topography is often plain~\cite{spekken2015,griffel2019,bolfe2020}. A \textit{productive crop field} is an area consistently used for the cycle of growth and harvest of a crop, typically yearly but more often in some regions with favorable soil and weather.

Thanks to the availability of a massive amount of labeled data, the development of artificial intelligence~(AI), particularly machine learning~(ML) and deep learning~(DL), has allowed for acceleration and improvement in many areas of agriculture~\cite{garcia2019,persello2019,masoud2020,waldner2020,waldner2021}. Despite all the advances, having an adequate amount of labeled data to train such methods can be costly and not always affordable for precision farming.

Usually, farmers rely on experts to build the field boundaries in dedicated Geographic Information Systems (GIS) software. 
Despite the expert's ability and knowledge, finding crop fields and drawing their boundaries has been a major, \replace{error-prone}{expensive}, and time-consuming challenge~\cite{wagner2020}. The larger the customer's land is, the more significant the number of fields to be created. 
For instance, there are farms in Brazil with more than eighteen thousand fields that require regular updates to reflect their accurate states. Finding productive fields is even more challenging because it requires previous knowledge of these areas or the analysis of satellite images over time.

Aiming to facilitate access to data ready to support ML and DL models for productive crop field detection, we introduce a new dataset of productive fields based on agriculture machine's operation combined with Sentinel-2 images collected over a period of time. Firstly, this information is aggregated in geospatial L12 \replace{hexes}{hexagons}, establishing a common ground for employing different satellite sources with different resolutions. Secondly, typical farmers are usually interested in monitoring productive areas rather than all possible fields since many are not fertile or suitable for any crop. For this purpose, the proposed dataset contains highly accurate positive samples (i.e., productive crop fields) and inferred negative ones, making it well-suited for positive and unlabeled learning~\cite{denis2005}. Moreover, recent advances suggest that self-supervised methods (e.g., contrastive learning) may also provide a promising alternative even for cases where labeled samples are scarce~\cite{guldenring2021}. Finally, we offer state-of-the-art ML and DL benchmark results for further comparison, including classification methods with different training strategies (i.e., supervised learning, self-supervised learning, and semi-supervised classification of unlabeled data).

\section{Related Work}

Automatically detecting productive crop fields is a challenge that intersects multiple areas, and its solution could fill important gaps in agriculture management and environment control~\citep{bolfe2020}. In this context, \citet{crommelinck2019} has worked on boundary detection of agricultural fields to provide automated recording of land rights, a slightly different purpose despite sharing common problems. However, despite the proven benefits of fully automated boundary detection and the research already accomplished, it is still considered an open problem with significant innovation opportunities, especially when sufficient training data is unavailable~\citep{waldner2021,yang2020}.

\citet{north2019} used classical computer vision techniques to detect the boundaries of farm fields in New Zealand. However, their technique resulted in only 59\% accuracy, which is insufficient for most applications. Moreover, their strategy is highly impacted by variations caused by the season in the images extracted. Still based on classic computer vision, \citet{evans2002} used a region-based technique called canonically-guided region growing to segment multi-spectral images and, despite being more accurate than previous work, it was applied to a very small and manually annotated set of images requiring further investigation in a more broad collection of samples.

Since classical computer vision techniques alone are often insufficient to achieve good results, further research evaluated ML models to detect crop boundaries. For instance,~\citet{garcia2017} achieved an accuracy of 92\% using an ensemble algorithm called RUSBoost~\citep{seiffert2010} to merge superpixels and group blocks of the image, which were part of the same field. This study demonstrated how ML is a promising alternative and a precursor to further development using DL~\citep{garcia2017}. \add{In the same year,~\citet{kussul2017} applied DL to segment and classify crop types. They demonstrated the advantages of deeper networks by comparing a Multilayer Perceptron~(MLP) with a CNN, where the latter achieved higher accuracies.} Later,~\citet{garcia2019} applied the U-Net, a convolutional neural network~(CNN) for semantic image segmentation, to perform the same task with improved performance~\cite{garcia2019,garcia2017,garcia2018}. 

\remove{In the same year, //citet|kussul2017| applied DL to segment and classify crop types. They demonstrated the advantages of deeper networks by comparing a Multilayer Perceptron~(MLP) with a CNN, where the latter achieved higher accuracies.}
Concurrently, \citet{persello2019} designed a DL approach using SegNet, a CNN commonly used for semantic segmentation in Very High Resolution~(VHR) images. \replace{In SegNet, only the pooling indices are transferred to the expansion path from the compression path using less memory. In contrast, in U-Net, entire feature maps are transferred from the compression path to the expansion one, demanding much more memory. Both techniques attained remarkable performance. However, }{In this work, }some significant mistakes were generally associated with poor-quality training data~\citep{persello2019}. The authors extend their work to medium-resolution images from Sentinel-2 \replace{using two different DL architectures, SRC-Net and MD-FCN. Ultimately, }{, however, }they observed a severe limitation during training since \replace{these}{the employed} methods comprised many convolutional layers resulting in a time-consuming task~\cite{persello2019,masoud2020}.

\citet{wagner2020} designed a modified version of the growing snakes active contour model based on graph theory concepts~\cite{wagner2020}. \remove{This graph-based growing contour technique can extract complex boundary networks in agricultural landscapes requiring little supervision. Evaluated on rural area maps, it detected 99\% of total acreage, but it achieved a poor performance close to urban areas.} In sequence, \citet{wagner2020.2} integrated a DL approach into their already proven graph-based model. \remove{For this, an MLP was used to make a pixel-wise prediction of whether it was a boundary.} The method achieved high results in rural areas but required further investigation of the missed boundaries, especially in areas closer to the cities~\cite{wagner2020.2,wagner2020}.

\citet{waldner2020} compared DL methods, starting with ResUnet-a and followed by an improvement named Fractal-ResUNet, a network designed for semantic segmentation of agricultural images. \remove{It employs hierarchical watershed segmentation, a method that relies on graph theory, suggesting this is a successful approach to address the challenge of segmenting cropland images.} The best accuracies were achieved with Sentinel-2 images, and it is considered state-of-the-art in boundary detection of agricultural fields. However, it depends on a large amount of training data with high-quality labeling to achieve consistent results~\cite{waldner2020,waldner2021}.

Most aforementioned studies were devised for detecting or segmenting all possible fields, regardless of whether they are productive. Only a few consider productive fields, but even these disregard their eventual changes over time. 

\section{Productive Fields Dataset}
The dataset is stored in tabular format, where each row corresponds to a hexagon, along with a timestamp and satellite band values. The hexagon format was used for geo analysis since it facilitates storing and modeling spatiotemporal data from multiple disparate sources while avoiding using rasters.  We used the H3 Python library\footnote{The H3 Python library is available at \url{https://pypi.org/project/h3/}. Access on May 4, 2023.} to create such hexagons with a level of L12. At this level, each hexagon has edges of approximately 10 meters, resulting in a hexagonal area that covers 307 square meters, slightly bigger than three Sentinel-2 pixels. This resolution offers a balanced trade-off between granularity and computational efficiency while training the models. Moreover, hexagons of this size can capture fine-grained variations in satellite data within a manageable dataset size, enabling efficient processing and analysis.

\begin{figure}[!htb]
    \centering
    \includegraphics[width=8cm]{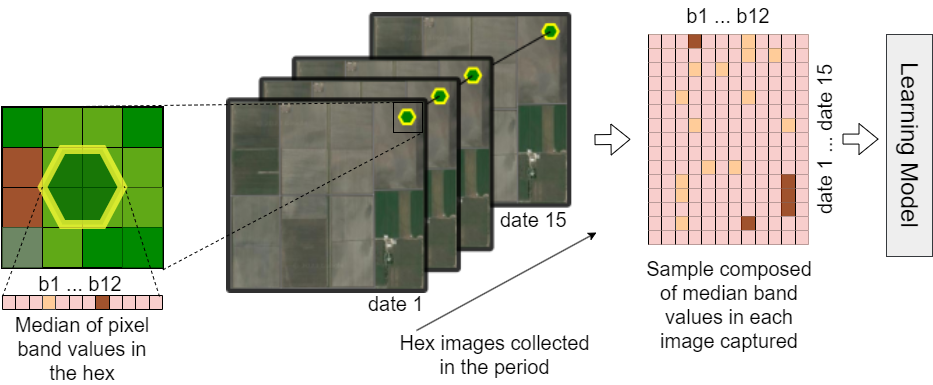}
    \caption{\label{figure:e2e}
\add{Sequence to build a sample based upon the hexagon time-series
}}
\end{figure}

The timestamps are associated with the date and time the Sentinel-2 images were captured, providing temporal information. There are images from 2018-10-29 17:04:21 to 2019-08-15 16:59:01. Finally, the twelve band values correspond to the pixel median values in the area covered by that particular H3 hexagon. With this, we can summarize the satellite data for each hexagon in a compact and interpretable format, facilitating downstream ML analyses. Figure~\ref{figure:e2e} illustrates how we create the samples used as the input for the learning models.

Table~\ref{table:dataset_description} summarizes the dataset and its fields. The dataset includes 17 productive corn crop fields in the US within the same geographical area. It has 106,735 rows, each representing a hexagon captured on a specific date and time, followed by the 12 Sentinel-2 band values. From these, 75,990 rows are labeled as positive (i.e., productive crop fields hexagons tracked over the period) and 30,745 as negative (i.e., non-productive crop fields hexagons tracked over the period). The amount of positive hexagons representing geolocations is 5,066, while the number of negative hexagons is 2,050 resulting in a slightly unbalanced dataset. The number of images for each hexagon collected on different dates varies from 15 to 38, and the averages for each field are represented in the last column.

\begin{table}[!htbp]
    \centering
    \caption{\label{table:dataset_description}Productive Fields Dataset Description} 
    \begin{tabular}{llllllll}
        \toprule
        \toprule
        \multirow{2}{*}{} & \multicolumn{6}{c}{Fields: 17}\\
        \cmidrule{2-7} \\
        \multirow{2}{*}{} & \multicolumn{2}{c}{Hexes (samples)} & \multicolumn{4}{c}{Hex images collected in the period} & \multicolumn{1}{c}{}\\
        \cmidrule(lr){2-3}  \cmidrule(lr){4-7}   \\
        \multicolumn{1}{c}{Field} & \multicolumn{1}{c}{positive} & \multicolumn{1}{c}{negative}& \multicolumn{1}{c}{positive} & \multicolumn{1}{c}{negative}& \multicolumn{1}{c}{average}\\
        \midrule
        F01&558&210&9,860&3,539&17.67 $\pm$0.48\\
        F02&734&151&14,242&2,863&19.40 $\pm$0.74\\
        F03&59&85&1,075&1,528&18.22 $\pm$0.65\\
        F04&350&150&8,326&3,518&23.79 $\pm$0.48\\
        F05&75&134&1,815&3,001&24.20 $\pm$0.90\\
        F06&288&60&4,999&1,044&17.36 $\pm$0.68\\
        F07&418&140&9,409&2,950&22.51 $\pm$0.70\\
        F08&425&154&8,584&3,084&20.20 $\pm$0.82\\
        F09&147&127&2,993&2,563&20.36 $\pm$0.71\\
        F10&369&176&6,973&3,194&18.90 $\pm$0.40\\
        F11&228&57&4,813&1,172&21.11 $\pm$0.59\\
        F12&91&96&1,701&1,738&18.69 $\pm$0.53\\
        F13&151&86&3,271&1,840&21.66 $\pm$0.49\\
        F14&275&216&4,871&3,825&17.71 $\pm$0.46\\
        F15&370&198&7,369&3,913&19.92 $\pm$0.66\\
        F16&271&25&5,030&413&18.56 $\pm$0.69\\
        F17&281&17&5,099&297&18.15 $\pm$0.97\\
        \bottomrule
        \bottomrule
    \end{tabular}
\end{table}

\remove{
begin|figure|[!htb]
Distribution of the number of samples by satellite image date.
end|figure|
}

\begin{figure}[!htbp]
    \centering
    \includegraphics[width=8cm]{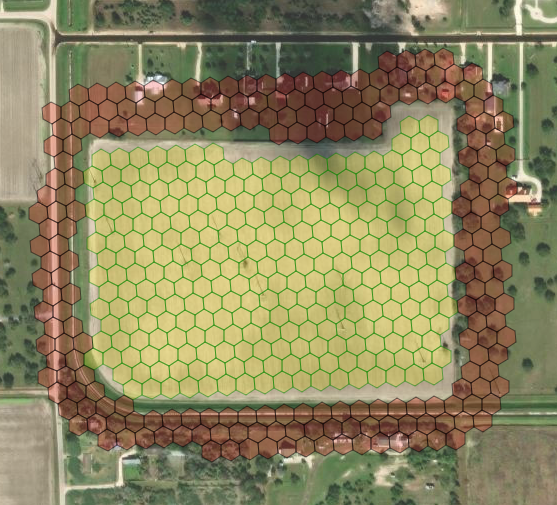}
    \centering\caption{\label{figure:field_hexes}Productive crop field filled with positively labeled \replace{hexes}{hexagons} in green surrounded by a three \replace{hexes}{hexagons} layer of inferred negative samples.}
\end{figure}

We have labeled each hexagon as positive or negative based on the productive agriculture machine operations, such as planting or harvesting. Specifically, hexagons containing evidence of these operations were labeled positive, while neighboring hexagons within a three-layer radius were labeled as negative, as shown in Figure~\ref{figure:field_hexes}. Labeling neighboring samples as negative is particularly interesting because they are often the hardest for prediction models to classify.

\begin{figure}[!htbp]
    \centering
    \includegraphics[width=8cm]{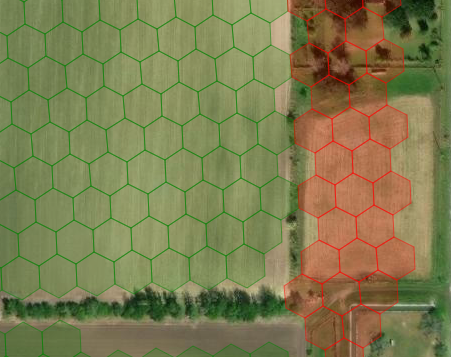}
    \centering\caption{\label{nfield_hexes}\add{A non-productive field properly classified with red \replace{hexes}{hexagons}}}
\end{figure}

\add{One noteworthy aspect of the proposed method is its ability to distinguish non-productive fields that share the same shape as productive ones. Even a highly skilled individual would struggle to make this distinction. \replace{The key reason behind this lies in the trained model's ability to recognize patterns in the Normalized Difference Vegetation Index (NDVI) over time rather than relying solely on static images}{The key reason behind this lies in the trained model’s ability to recognize patterns of band values over time from non-productive areas as distinctly different than productive cropped areas}. This approach leads to significantly more accurate results. Figure~\ref{nfield_hexes} shows an example of a non-productive field correctly identified.}

Some false negatives may occur where productive fields are not detected due to the absence of agriculture machines equipped with satellite receptors or, for many reasons, when data are not shared. However, the dataset was manually curated to minimize the number of false negatives, specifically targeting the hexagonal regions between crop and non-crop fields. Despite these efforts, some false negatives may still be present. Even so, the proposed dataset is a valuable resource for future remote sensing research since, to our knowledge, it is the first to offer high-quality labeled productive crop fields.

\section{Benchmark Results}
This section offers state-of-the-art ML and DL benchmark results to facilitate further comparison. We have analyzed the semi-supervised classification of unlabeled data and deep neural networks with different training strategies (i.e., supervised and self-supervised learning). We can use these models to predict previously unseen \replace{hexes}{hexagons} or pixels based on the precision required for the output image. Both the dataset and implementation notebooks are publicly available at \href{https://github.com/egnascimento/productivefieldsdetection}{https://github.com/egnascimento/productivefieldsdetection}.

\subsection{Data processing}
\begin{figure}[ht]
    \centering
    \includegraphics[width=8cm]{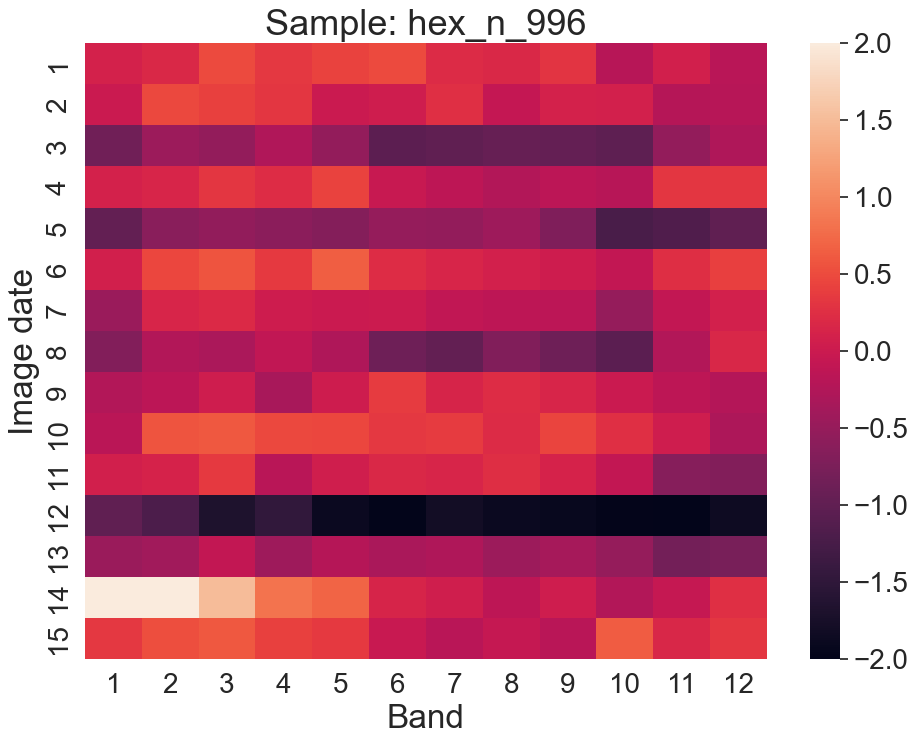}
    \caption{\label{figure:sample}Multitemporal time series sample visualization with 15 rows (i.e., each row corresponding to a unique date a satellite image was captured of the \replace{hex}{hexagon} location) and 12 columns (i.e., satellite band values.}
\end{figure}

We processed and grouped the data to construct bidimensional time series samples. Each comprises 15 randomly selected image dates, as shown in Figure~\ref{figure:sample}.  A total of 8,342 grouped multitemporal time series samples resulted. Selecting sparse dates throughout the year is the best strategy for a comprehensive collection comprising different weather seasons and crop stages.


The original dataset was split into 80\% training, 10\% test, and 10\% validation data keeping the balance between positive and negative labeled data.

\subsection{Machine learning and deep learning methods}
Based on the dataset's characteristics and training strategies, three state-of-the-art methods on ML and DL were chosen:

\begin{itemize}
\item \textit{Positive Unlabeled (PU)}~\citep{denis2005} is a semi-supervised classification approach of unlabeled data that is particularly suitable for the current study, given the availability of accurate positive data, the lack of negative one, and the ease of obtaining unlabeled data. PU learning relies on supervised methods, and we selected Support Vector Machines (SVM) for this evaluation.
\item \textit{Triplet Loss Siamese network}~\citep{schroff2015} is designed to learn representations that effectively discriminate similar and dissimilar samples in a latent space. The Triplet Loss function compares the distance between an anchor sample and a positive one against the distance between the anchor sample and a negative one. 
\item \textit{Contrastive Learning}~\citep{ashish2020} is especially recommended in scenarios of limited data availability (e.g., regions with precision agriculture limited or nonexistent). Contrastive learning methods are employed to address the shortage of samples by employing data augmentation. This approach has been widely used to tackle diverse issues with significant potential for detecting crop fields. We have implemented the well-known SimCLR~\citep{chen2020}. As Contrastive Learning is highly dependent on good data augmentation, a random jitter of up to $\pm$10\% was applied to all bands to create augmented samples.
\end{itemize}

\begin{figure}[ht]
    \centering
    \includegraphics[width=8cm]{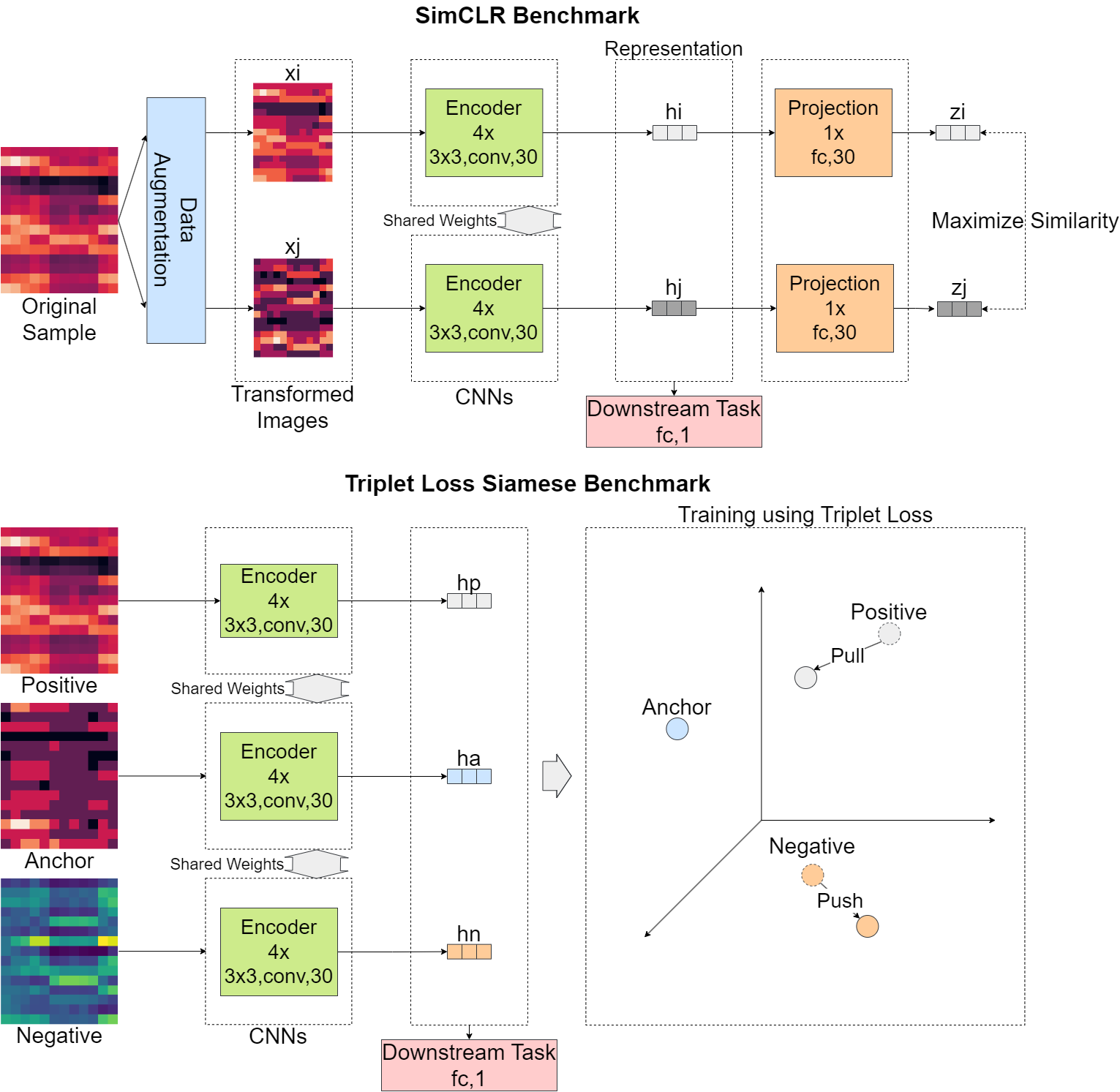}
    \caption{\label{figure:architecture}\add{Neural network architectures employed in the benchmark testing}}
\end{figure}

We have implemented the DL-based neural networks using 4 convolutional layers with a kernel size of 3 and stride of 1, as demonstrated in Figure~\ref{figure:architecture}.

\subsection{Results}
Table~\ref{table:metrics} presents the results, including metrics suitable for evaluating this problem~\citep{zhang2018}. The accuracies of PU Learning, Contrastive Learning, and Triplet Loss Siamese were 95.36\%, 97.05\%, and 97.47\%, respectively. PU learning achieved accuracies slightly below DL methods, a remarkable performance that can still be significantly improved by changing the underlying method to a DL neural network. The F1 score weighted average was used since the dataset is imbalanced. Additionally, we included individual F1 scores by class to demonstrate that the best results came from the more reliable samples labeled by machine data. Finally, the analysis of the Matthews correlation coefficient (MCC), appropriate to imbalanced  datasets, supported the conclusion that the Triplet Loss Siamese method achieved the best overall results.

\begin{table}[htbp]
    \centering
    \caption{\label{table:metrics}Benchmark results}
    \begin{tabular}{llllll}
        \toprule
        \multirow{2}{*}{Models} & \multicolumn{5}{c}{Metrics}\\
        \cmidrule{2-5} \cmidrule{6-6} \\
        {} & Accuracy & F1-weighted  & F1-Neg& F1-Pos & MCC\\
        \midrule
        DL Baseline & 96.62\%& 96.61\%& 94.26\%& 97.61\%& 0.92\\
        PU SVM & 95.36\%& 95.36\%& 92.27\%& 96.68\%& 0.89\\
        C. Learning & 97.05\%& 97.03\%& 94.96\%& 97.91\%& 0.93\\
        Triplet Loss & \textbf{97.47\%}& \textbf{97.47\%}& \textbf{95.77\%}& \textbf{98.19\%}& \textbf{0.94}\\
        \bottomrule
    \end{tabular}
\end{table}

\begin{figure}[ht]
    \centering
    \includegraphics[width=8cm]{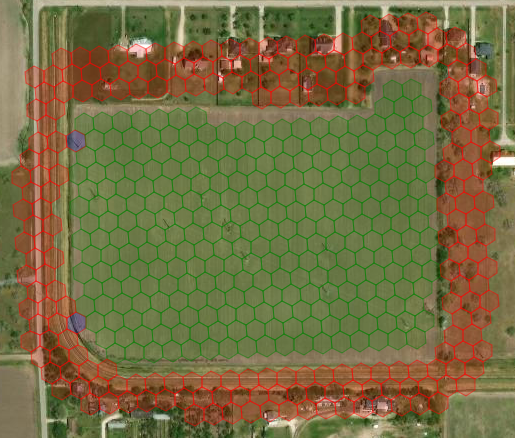}
    \caption{\label{figure:predicton_mistakes}Blue hexagons represent prediction mistakes, usually found between productive and non-field areas.}
\end{figure}

Analyzing the results, we have observed that all models missed, in most cases, the prediction of samples located between field and non-field areas, as presented in Figure~\ref{figure:predicton_mistakes}.  These mistakes, however, should not be relevant when performing classification at the pixel level. 

\section{Conclusion}
The dataset proposed in this paper represents an important alternative to training highly accurate prediction models that automatically detect productive crop fields. The main advantage is that it does not rely on manual work, which ensures, although not perfect, high-quality labeled samples in large volume since they are being continuously produced by machines equipped with precision agriculture devices. In sequence, we evaluated three different and complementary state-of-the-art ML and DL algorithms well-suited to the problem of detecting productive crop fields.
The benchmark results present a performance high enough for most real applications involving detecting and delineating productive crop fields.
For future work, we suggest collecting more data, including different crops. We also suggest improvements in the data augmentation when contrastive learning is used and evaluating an instance segmentation algorithm, possibly a graph-based one, which should be connected to the end of the pipeline to produce shape files ready-to-use by precision agriculture systems.

\section*{Acknowledgment}

We would like to express our gratitude to John Deere for generously sharing the data. This partnership has reinforced the company's unwavering commitment to advancing scientific research while prioritizing the confidentiality and security of its customers' data. We deeply appreciate this collaboration and recognize the crucial role that technology companies like John Deere play in supporting scientific innovation.

\ifCLASSOPTIONcaptionsoff
  \newpage
\fi



%
\footnotesize
\bibliographystyle{IEEEtranN}
\bibliography{bibtex/bib/IEEEabrv,bibtex/bib/references}

\end{document}